\documentclass[letterpaper, 10 pt, conference]{ieeeconf}  

\IEEEoverridecommandlockouts                              

\usepackage{float}
\usepackage{graphicx} 
\usepackage{amsmath} 
\usepackage{bm}    
\usepackage{amssymb}  
\usepackage{caption}
\usepackage{subcaption}
\usepackage{float}
\usepackage[ruled,vlined]{algorithm2e}

\usepackage{color}
\usepackage[bookmarks=false]{hyperref}
\usepackage[noadjust]{cite}

\DeclareMathOperator*{\argmax}{argmax} 

\newcommand{\R}{\mathbb{R}}

\newcommand{\unitarypolicy}{$\pi^u_{\theta}(a \vert s)$}
\newcommand{\statenet}{$T_s = \rho_{\phi}(f)$} 
\newcommand{\actionnet}{$T_a = \rho_{\psi}(f,\bar{s}_u)$} 

\title{\LARGE \bf
Learning Generalizable Pivoting Skills
}

\author{Xiang Zhang$^{1}$, Siddarth Jain$^{2}$, Baichuan Huang$^{3}$, Masayoshi Tomizuka$^{1}$, and Diego Romeres$^{2}$
\thanks{$^{1}$Mechanical Systems Control Lab, UC Berkeley, Berkeley, CA, USA.
        {\tt\small \{xiang\_zhang\_98, tomizuka\}@berkeley.edu}}%
\thanks{$^{2}$Mitsubishi Electric Research Laboratories (MERL), Cambridge, MA,
USA 
        {\tt\small \{sjain,romeres@merl\}.com }}%
        \thanks{$^{3}$Department of Computer Science, Rutgers University, Piscataway, NJ, USA.
        {\tt\small baichuan.huang@rutgers.edu}}%
}

\begin{document}

\maketitle
\thispagestyle{empty}
\pagestyle{empty}

\begin{abstract}
The skill of pivoting an object with a robotic system is challenging for the external forces that act on the system, mainly given by contact interaction. The complexity increases when the same skills are required to generalize across different objects. This paper proposes a framework for learning robust and generalizable pivoting skills, which consists of three steps. First, we learn a pivoting policy on an ``unitary'' object using Reinforcement Learning (RL). Then, we obtain the object's feature space by supervised learning to encode the kinematic properties of arbitrary objects. Finally, to adapt the unitary policy to multiple objects, we learn data-driven projections based on the object features to adjust the state and action space of the new pivoting task. The proposed approach is entirely trained in simulation. It requires only one depth image of the object and can zero-shot transfer to real-world objects. We demonstrate robustness to sim-to-real transfer and generalization to multiple objects.

\end{abstract}
%
\section{INTRODUCTION}

\begin{figure}
    \centering
    \includegraphics[width=230pt]{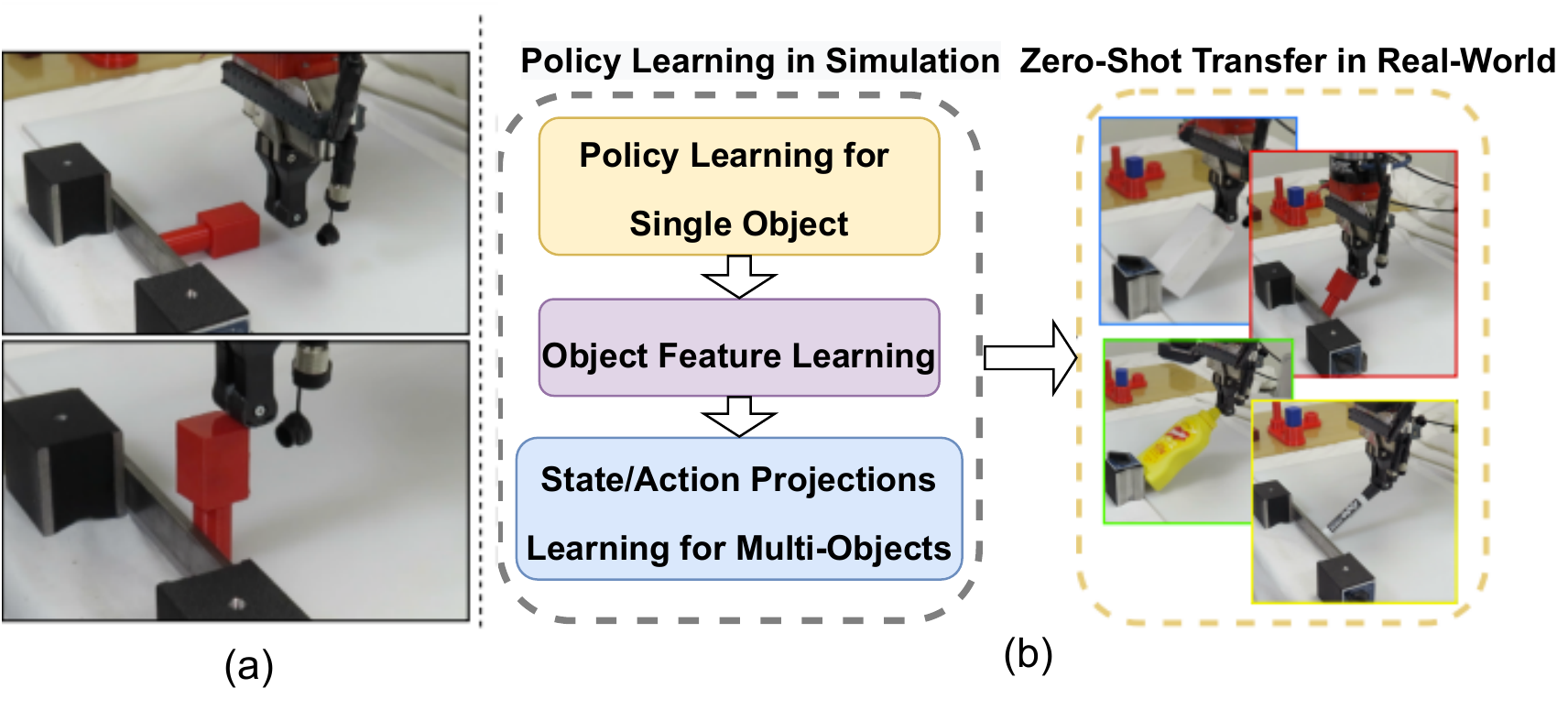}
    \caption{a) Pivoting task setup. b) policy learning in simulation and zero-shot transfer to real-world}
    \label{fig:overview}
\vspace{-0.5cm}
\end{figure}

Table-top manipulation skills like pivoting are required to reorient objects often to create pre-conditions for other manipulation skills. For example, a book on the table may be too large for a robot to grasp, and a peg may lay in the wrong orientation for an insertion task. However, reorienting the book and the peg with a pivoting motion creates the conditions for a successful grasp. Fig.~\ref{fig:overview}(a) depicts the pivoting setup when the object is between two external surfaces, and the robot needs to exploit the interaction with these surfaces to pivot the object. A significant difficulty for pivoting is that the robot must maintain the object-gripper and object-surfaces contacts. Furthermore, multiple objects' different kinematic and inertial properties entail additional complexity like instability, slipping, and rolling properties.

This paper proposes a framework for learning a generalizable robotic skill of pivoting real-world objects from only simulation experience. Specifically, we would like to adapt the pivoting policy on one object to multi-objects based on the object depth image. As shown in Fig~\ref{fig:overview}(b), the proposed framework consists of three parts. First, a pivoting policy is learned using RL to pivot one specific object, which we call the ``unitary'' object. Then, to extract low-dimensional object information from the high-dimensional object depth images, we employed supervised learning on a dataset collected in simulation to learn a feature space by predicting the object class and size. Finally, object-specific state and action projections are learned to adapt the unitary policy to a novel object by adjusting the state and action space. These projects are linear transformations obtained from the object features and learned with a policy-gradient based approach. Intuitively, the state projections adjust the states of the new object to make it similar to the unitary object. Accordingly to the new object, the action projections alter policy outputs to improve policy performance. The proposed approach is trained entirely in the simulation and zero-shot transferred to a series of real-world pivoting tasks.

In summary, our work makes following contributions:
\begin{itemize}
  \item It introduces an RL framework to learn generalizable pivoting robotic skills of real-world objects with training only in simulation. 
  \item It proposes a policy-gradient based approach to learning state/action transformations to achieve generalization to unseen objects.
  \item It provides an extensive evaluation of our proposed approach in both simulation and real-world experiments with promising success rates.
\end{itemize}

\section{RELATED WORKS}
\subsection{Robot Contact-Rich Manipulation}

Approaches for robot contact-rich manipulation can be mainly categorized into two classes: model-based methods and model-free methods. Model-based approaches develop open-loop and feedback control strategies by modeling contact dynamics. However, the contact dynamics are hybrid because of different contact modes during contact. Researchers either model the hybrid contact dynamics as complementarity constraints \cite{Yuki2022Robbust,Jin2021Trajectory,raghunathan2022pyrobocop} or directly formulates the mixed-integer programming problem \cite{shirai2022chance} to obtain the optimal control law. Recently, some approaches \cite{hogan2018reactive, cheng2022contact} utilized the hierarchical framework to solve the discrete contact modes and robot motions separately to reduce the calculation time. However, model-based approaches require strong assumptions in the initial conditions and contact pairs as well as system identification to obtain the physical parameters of the objects.

Unlike model-based approaches, model-free approaches skip the complex contact dynamics modeling and directly learn a policy to achieve the manipulation task. Recently, reinforcement learning (RL) methods have been introduced to learn manipulation skills by optimizing a designed reward. Examples can be found in robot assembly tasks \cite{zhang2022learning, zhao2022offline, zhang2021learning}, bi-manual manipulation \cite{chitnis2020efficient}, and tabletop manipulation \cite{nasiriany2022augmenting}. Specifically, authors of \cite{zhou2022learning} investigated pivoting tasks using RL. However, their pivoting policy is learned on one object during training and can only apply to similar-sized objects, limiting their application.
\subsection{Learning Generalizable Robot Skills}
Generalization is a significant concern for robot skill learning because we are interested in robot skills that are robust to environments and task condition changes. Indeed, different objects have different geometric shapes, physical properties, and contact dynamics. Each of these differences would change how the robot interacts with the object and make the generalization of skills difficult. Researchers have proposed approaches to learn robust skills that work for different task settings or can adapt quickly to new tasks. Domain randomization \cite{laskin2020reinforcement} can be applied for robustness to force the learned policy to extract useful information from the state. Furthermore, Model Agnostic Meta Learning (MAML) \cite{finn2017model} and online model learning \cite{offlineWang2022} can adapt the dynamics model or policy to new tasks within a few trials.

Another approach is to encode the task information into the policy. Thus, the robot actions generated by the learned policy are conditioned on the tasks. This way, the robot adjusts skills according to different tasks to achieve generalization. Commonly, the task information can be inferred from either robot trajectories, images, or one-hot encoding. This approach has also proven successful on assembly \cite{zhao2022offline}, pick and place \cite{hausman2018learning} and legged-locomotion \cite{li2020learning,rakelly2019efficient}.

Researchers also consider skill generalization as a domain adaptation problem. For different tasks, the state and action space may change according to the task settings (e.g., robot type, object shape, goal states, etc.), and the analysis is needed to transfer a learned policy to other tasks. One approach is finding a shared latent space between tasks invariant to task settings. Once the skill is learned in the invariant latent space, it can be transferred to different task settings using task-specific mappings \cite{yin2022cross,kim2020domain}. Other methods applied direct mapping on the state or action space to transfer the skill in the source domain to the test domain \cite{taylor2007transfer,Tang2016Robotic}. However, for previous approaches, the state or action space mappings are either obtained manually by analyzing the difference between spaces or obtained by the point cloud registration algorithm. Therefore, prior human knowledge is utilized to find out the correct mapping. Our proposed method automatically discovers the underlying mappings on state or action space by maximizing the trajectory return.

\section{PROBLEM FORMULATION}\label{sec:problem_formulation}
In this work, we propose a framework to learn the robotic skill of pivoting real-world objects in a structured environment with zero-shot transfer learning from simulation to the real world. The main focus is generalizing the learned pivoting skill to arbitrary unseen objects.

Consider an environment as represented in Fig~\ref{fig:overview}(a), where a rigid object $o$ is at rest on a flat surface like a table, and it is in the proximity of a second surface, called wall, perpendicular to the table. The object $o$ can be manipulated by any end-effector of a robotic manipulator that can establish a patch contact with the object, like the fingers of a gripper. The goal is to learn a policy, $\pi(\cdot \vert s)$ where $s$ is the system state, that can first bring the gripper in contact with $o$, then establish contacts between $o$ and the wall, and finally utilize the environmental contacts gripper-object, object-table and object-wall to pivot the object to a stand-up position. 

Notice that we do not fix the initial condition of any element in the environment. In particular, we do not require the wall-object and the gripper-object contacts to preexist, and we do not assume to know the exact wall and object position and orientation.

The questions we try to answer in this scenario are:
\begin{enumerate}
    \item How do we learn a zero-shot pivoting policy that transfers from simulation to the real world?
    \item Can the same policy generalize to different objects?
\end{enumerate}

\begin{figure*}
    \centering
    \includegraphics[width=380pt]{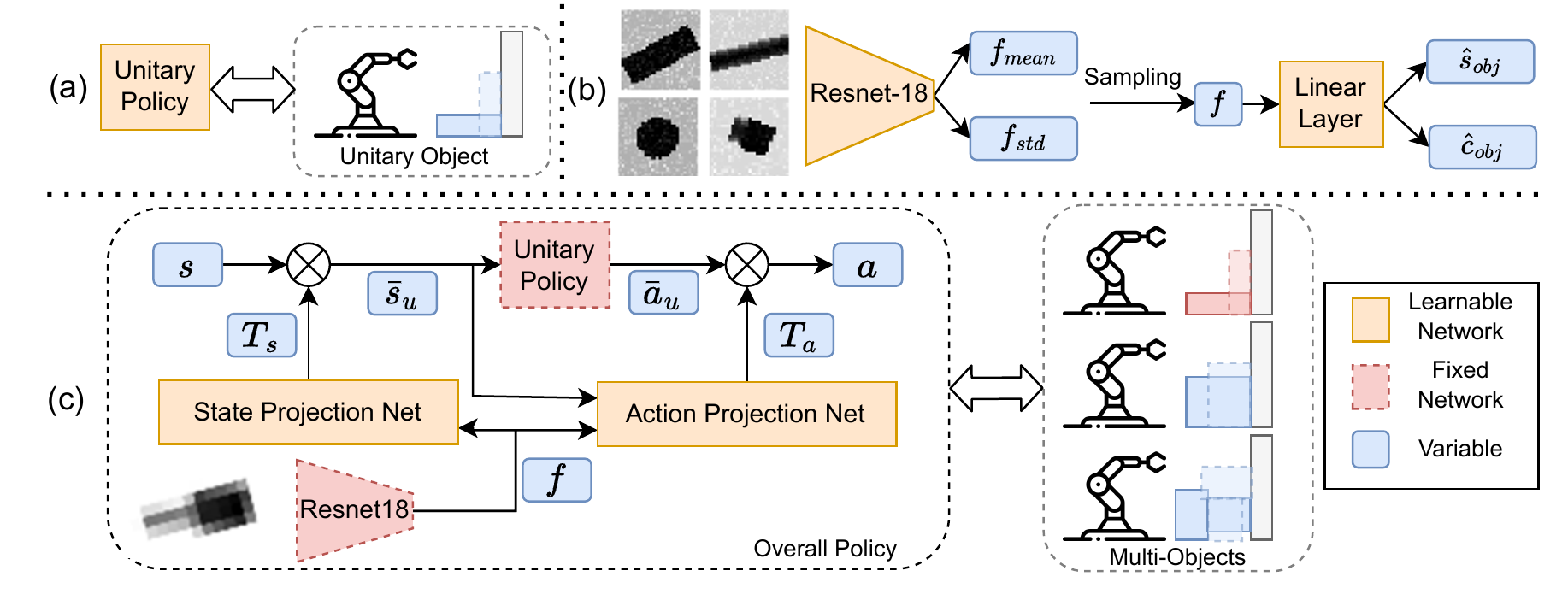}
    \caption{Three steps of pivoting policy learning in simulation: a) learning pivoting policy on the unitary object. b) object feature learning by predicting object class and size c) state/action projection nets learning on multi-objects pivoting}
    \label{fig:simulation_learning}
\end{figure*}
\section{PROPOSED APPROACH}\label{sec:approach}
In this section, we introduce our proposed approach to solve the problem described in Section~\ref{sec:problem_formulation}.
The basic idea is that the pivoting operation of different objects might be computed as a transformation of a policy learned on one primary object rather than be learned from scratch every time. We decompose the approach into three main steps. First, a pivoting policy is learned in simulation for an arbitrary object, denominated the \emph{``Unitary''} object. Second, a latent space of object features is learned to represent the shape of objects based on synthetic depth images generated in simulation. Finally, two neural networks are trained to adapt the unitary policy to a novel object by adjusting the state and action space. The overall framework is shown in Fig~\ref{fig:simulation_learning}, and the steps are detailed in the following section. 

\subsection{Reinforcement Learning for Pivoting the Unitary Object} \label{subsec:RL_for_single}
\begin{figure}
    \centering
    \includegraphics[width=160pt]{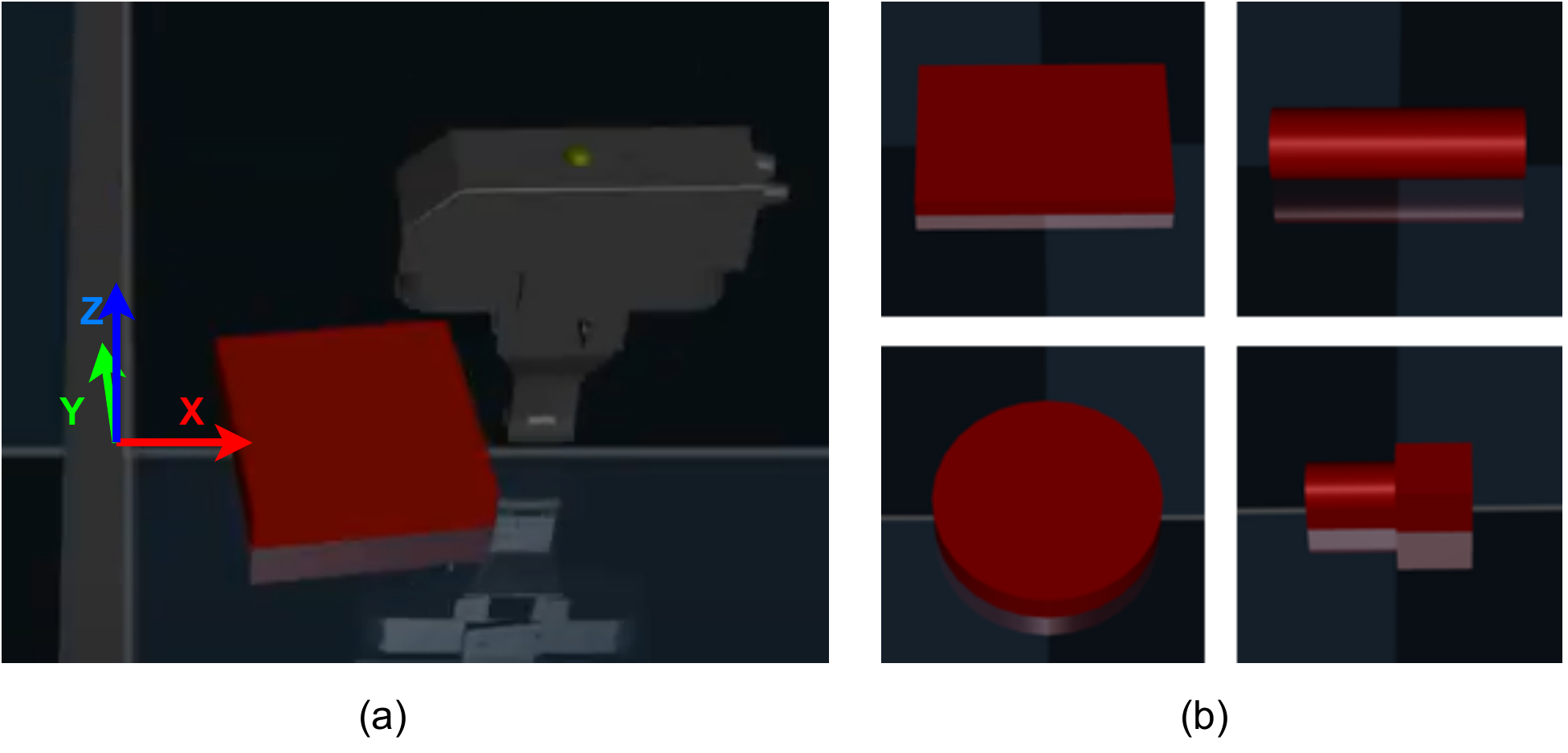}
    \caption{Simulation in Mujoco: a) training on the unitary object, b) four classes of objects (box, circle, cylinder, peg)}
    \label{fig:simulation}
\vspace{-0.5cm}
\end{figure}
This section describes the RL framework to learn the policy to pivot the unitary object. A RL framework is mathematically defined by an $MDP:=\{S, A, R, T, \gamma \}$ which is a tuple where $S$ is a set of states, $A$ is a set of actions, $R$ is a reward function that assigns a real value to each state/action pair, and $T$ is the state-transition function and $\gamma$ is the discount factor. 

\noindent \textbf{The simulation environment.} The unitary pivoting policy \unitarypolicy~parameterized by $\theta$ is learned in a Mujoco \cite{todorov2012mujoco} simulation environment, as shown in Fig.~\ref{fig:simulation}(a). The simulation includes the robot gripper and the unitary object, which is a $9\times9\times3~{cm}^3$ box. The dimensions are arbitrarily chosen, and they don't affect the algorithm. Moreover, a rigid wall is placed at what we consider the world frame origin to act as an external surface.\\ 
The details to train \unitarypolicy\ are as follows:

\noindent \textbf{The state space} is defined by three components: object pose $s_o$, gripper pose $s_g$, and the external forces measured by the F/T sensor at the wrist of the robotic manipulator, right above the gripper, $s_{F}$. The object and gripper pose include the cartesian position (X, Y, Z axes) and orientation in quaternion, $s_o, s_g \in  \R^7$ while $s_{F}$ contains the forces measured along the X, Y, and Z axes, $s_{F}\in\R^3$. Thus, the state $s:=[s_o, s_g, s_{F}] \in \R^{17}$. The maximum forces applied by the robot in the simulation are $\pm 10N$ in each axes, and $s_F$ is normalized to $\pm 1N$.

\noindent \textbf{The action space} is defined by the linear velocity of the robot gripper in X, Y, and Z axes as well as the angular velocity in the pitch direction, $a = [a_x,a_y,a_z,a_\rho]\in \R^4$. The actions are limited by a moving threshold set to $25~mm$. If the gripper moves more than this limit during training, the robot stops and proceeds to the next action.

\noindent \textbf{The reward function} is the distance between the current object rotation matrix $R$ and the goal rotation matrix $R^{goal}$ which is defined as:
\begin{equation}
    r = \frac{\pi}{2} - d, \text{ with } d = \arccos{\left(0.5(\text{Tr}(R^{goal}R^T)-1)\right)}
\end{equation}
where $Tr(\cdot)$ indicates the trace of a matrix. $\frac{\pi}{2}$ is added to the reward to make the initial reward close to $0$. 
This reward encourages the robot to pivot the object to the goal orientation $R^{goal}$, which is set as the orientation when the object is perpendicular to the ground.

\noindent \textbf{Domain randomization} is employed to improve the robustness of the pivoting policy with three kinds of noises:

1) Uncertainty of the wall position. The origin of the world frame is set at the wall, and we assume the exact position of the wall is unknown. Since the positions of the object and gripper are measured w.r.t the wall, we model this uncertainty by adding a zero-mean Gaussian noise with an std of 2 $cm$ to the object and the gripper positions.

2) Force measurement noise. To model the noise of the F/T sensor measurements, a zero-mean Gaussian noise with an std of $0.5~N$ is added to the measured force in the simulation.

3) Initial pose distribution. During the training, we randomized the initial pose of the object. Specifically, the object is placed with an offset to the wall and is oriented by a random angle. The initial offset $\Delta_x$ is sampled uniformly from range $[0, 5~cm]$. The initial rotation angle limit is defined as $\pm \arctan(\Delta_x/4.5)$, and the initial rotation angle is sampled from this range to avoid an unfeasible initial pose. 

\subsection{Representation Learning for Object Features} \label{subsec:representation}
The objective is to adapt the pivoting unitary policy to work with multiple objects. The kinematic information of the novel object is required to achieve this goal. We rely on representation learning to learn a low-dimensional feature space of the object based on top-down depth images of the objects. The RGB textures are superfluous for this task, and we do not rely on physical measurements because this would require an extra engineering step for the user. 

We first build a dataset of depth images in simulation to learn such a feature space. As depicted in Fig~\ref{fig:simulation}(b), we generated the dataset from four object classes $c_{obj}$: rectangular box, circle, cylinder, and peg.  
For each class, 100 randomly sized objects are generated. The generated dataset is composed as $\mathcal{D}:=\{\mathcal{I}_i,c_{obj,i},s_{obj,i}\}_i^{400}$ where $\mathcal{I}_i$ are the depth images of each object and $s_{obj,i} = [l_x, l_y, l_z]$ are the sizes, recorded as labels. Then, $\mathcal{D}$ is augmented ten times to 4000 data points by applying random translations, rotations, and Gaussian noise to each original data point.

Fig~\ref{fig:simulation_learning}(b) shows the proposed network $F(f \vert \mathcal{I})$ to learn the object features $f$ based on the object depth image $\mathcal{I}$. First, the standard Resnet18 architecture \cite{he2016deep} processes the object depth image $\mathcal{I}$ and outputs the mean, and the standard deviation of the object features $f_{mean}$ and $f_{std}$. Second, similarly to the variational auto-encoder~\cite{kingma2013auto}, we use the reparametrization trick to sample the object feature $f$. Finally, another linear layer outputs the predicted object size $\hat{s}_{obj}$ and logits for the object class $\hat{c}_{obj}$. The loss function to train the network is designed as follows:
\begin{align*}
    L &= L_{shape} + L_{class} + \beta L_{KL} = \lVert s_{obj} - \hat{s}_{obj} \rVert^2\\
      & + L_{CE}(c_{obj},\hat{c}_{obj}) + \beta  D_{KL}(N(f_{mean},f_{std}), N(0,1))
\end{align*}
where $L_{CE}$ is the cross entropy loss and $\beta$ is a weight on the KL divergence loss. The first two terms of the loss are for supervised learning to predict the object size and class. The KL divergence loss regulates the learned feature space and mitigates over-fitting as mentioned in \cite{kingma2013auto}.
\subsection{State and Action Projection Nets}\label{subsec:state_action_nets}
Standard domain randomization techniques such as the one used in Section~\ref{subsec:RL_for_single} do not generalize to objects with significantly different kinematic properties; reaching this level of generalization is one of our main goals.
The same MDP describes different objects, but the unitary object would get as input unseen states and output actions unsuited for multiple objects. However, as shown in Fig~\ref{fig:LC}(b), the trajectories of different objects during pivoting are similar in shape and possibly can be described by trivial transformations in the right space.

\begin{figure}
    \centering
    \includegraphics[width=220pt]{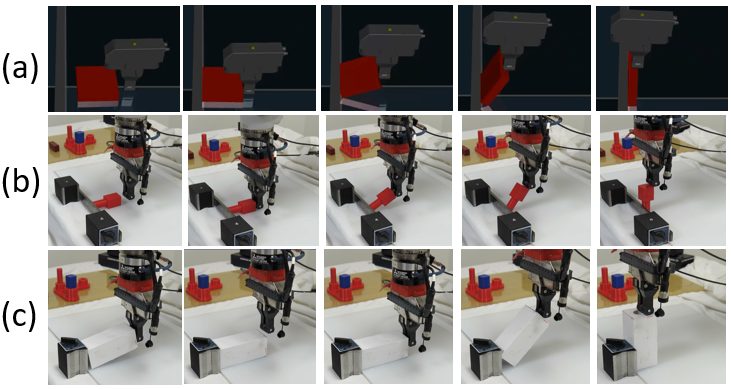}
    \caption{Snapshots of: a) pivoting the unitary object in simulation b) pivoting a peg in real-world c) recovery behavior.}
    \label{fig:snapshots}
\end{figure}

Inspired by this intuition, we propose learning object-based transformations to adjust the state and action spaces and generalize the unitary policy to novel objects instead of learning from scratch. Specifically, we choose to use linear transformations for simplicity and call these transformations \emph{state} and \emph{action projection nets} as depicted in Fig~\ref{fig:simulation_learning}(c). The \emph{State Projection Net}: \statenet\, is parameterized by $\phi$, and takes as input the feature of the object $f$ to output a diagonal matrix $T_{s}$ of state dimensions. The output is used as a linear operator to project the object state, $s$, to a space similar to the unitary state $\bar{s}_u$: $\bar{s}_u = T_{s} s$. The projected state $\bar{s}_u$ is fed into the unitary policy $\bar{a}_u = \pi^u_{\theta}(\bar{a}_u|\bar{s}_u)$ to generate the pivoting action. However, $\bar{a}_u$ needs to be transformed to work into the original object. For this reason, we train the \emph{Action Transformation Net}: \actionnet\,, that given $f, s_u$ outputs a diagonal matrix $T_{a}$ of action dimensions. That is used to compute the pivoting actions: $a = T_a\bar{a}_u$.

The overall action inference process can be summarized as:
\begin{align}
    a = T_a \bar{a}_u = T_{a} \pi^u_{\theta}(\bar{a}_u|\bar{s}_u)= T_{a}\pi^u_{\theta}(\bar{a}_u|T_{s} s)
    \label{eqn:infer_action}
\end{align}

Thus, the overall generalizable pivoting policy consists of: the unitary policy, the object feature extraction network, and the state/action projection nets. The former two are already trained, and the weights are frozen. Only the state/action projection nets need to be trained to adapt the pivoting policy to different objects. In particular, we use a policy gradient approach to learn the state/action projection nets to maximize the trajectory return of pivoting different objects. Suppose we collected a pivoting trajectory 
$\tau = (s_1, a_1, T_{s}, T_a,f, r_1,\dots, s_T, a_T, T_{s}, T_a,f,r_T)$ of an object with feature $f$, the advantage functions of the state/action transformation nets are, respectively:
\begin{equation}
    \hat{A}_s = \frac{1}{T}\sum^T_{i=0} \gamma^i r_i; \hat{A}_{at} = \sum^T_{i=t} \gamma^i r_i
    \label{eqn:advtange}
\end{equation}
where $r_i$ is the reward at time $i$ and $\gamma$ is the discount factor.
Then a simpler version of PPO \cite{schulman2017proximal} update is applied for both state and action projection nets:
\begin{align}
    &L_s(f, T_s,\phi, \phi_{old})= \\
    & \min \left(\frac{\rho_{\phi}(f)}{\rho_{\phi_{old}}(f)}, \text{clip}\left(\frac{\rho_{\phi}(f)}{\rho_{\phi_{old}}(f)}, 1+\epsilon_s, 1-\epsilon_s\right)\right)A_s \notag\\
    &L_a(f, s_u, T_a,\psi, \psi_{old})= \\
    & \min \left(\frac{\rho_{\psi}(f, s_u)}{\rho_{\psi_{old}}(f, s_u)}, \text{clip}\left(\frac{\rho_{\psi}(f, s_u)}{\rho_{\psi_{old}}(f, s_u)}, 1+\epsilon_a, 1-\epsilon_a\right)\right)A_a \notag
\end{align}

where $\frac{\rho_{1}(\cdot)}{\rho_{2}(\cdot)}$ is the ratio of likelihood of two projections and 
$\epsilon_s,\epsilon_a$ are clipping factors for update. The weights of the two projection nets are updated by:
\begin{align}
    \phi_{k+1} &= \argmax_{\phi} \mathop{\mathbb{E}}_{(f,T_s) \sim \pi_{\phi_k}} L_s(f, T_s,\phi, \phi_{k}) \label{eqn:state_update}\\
    \psi_{k+1} &= \argmax_{\psi} \mathop{\mathbb{E}}_{(f, s_u, T_s) \sim \pi_{\psi_k}} L_a(f, s_u, T_a,\psi, \psi_{k})
    \label{eqn:action_update}
\end{align}
Essentially, we maximize the likelihood of state or action transformations with higher reward-to-go. The details of our proposed approach are summarised in Algorithm~\ref{algo}. 
\begin{algorithm}
\SetAlgoLined
Initialize state and action projection nets $\rho_{\phi}(f), \rho_{\psi}(f, s_u)$ with random weights $\phi,\psi$

Initialize the unitary policy $\pi^u_{\theta}(\bar{a}_u|\bar{s}_u)$ and feature extraction network $F(f|I)$ with pretrained weights.

\For{i = 0, 1, 2, \dots until convergence}{
\For{iteration k = 1 to K}{
    Randomly sample an object $o_k$ with image $\mathcal{I}_k$
    
    Infer the object feature $f_k \thicksim F(f|\mathcal{I}_k)$
    
    Infer actions using~(\ref{eqn:infer_action}) to collect a trajectory
}

Calculate advantages for state and action transformations using~(\ref{eqn:advtange})

\For{iteration m = 1 to M}{
Update \emph{State/Action Projection Nets}~\eqref{eqn:state_update},\eqref{eqn:action_update}
}
}
 \caption{Learning state and action projections}
 \label{algo}
\end{algorithm}
\section{EXPERIMENTS}
\subsection{Simulation: Training and Validation Experiments}

\textbf{RL for Pivoting Unitary Object:} The pivoting policy for the unitary object, \unitarypolicy, is trained on MDP described in Section~\ref{subsec:RL_for_single} with the SAC \cite{haarnoja2018soft} algorithm using the implementation from RLkit \cite{rail-berkeley}. The policy and Q-function networks are parameterized as two-layer Relu networks with 128 and 256 units, respectively. The batch size is 1024, and learning rates are $5e-3$ for the Q-function and $3e-4$ for the policy. Fig~\ref{fig:snapshots}(a) depicts a sequence of snapshots of pivoting the unitary object in simulation after training the policy. The gripper first pushes the object towards the wall to establish contact between the object and the wall. Then, the robot moves upwards while pushing the object against the wall to start the pivoting. After the object rotates over a certain angle, we notice that the robot applies downward forces to maintain contact with the object, which is a robust way to stabilize the object and prevent the object from dropping off. Finally, the object is flipped up and standing on the table. 

\begin{figure}
    \centering
    \includegraphics[width=240pt]{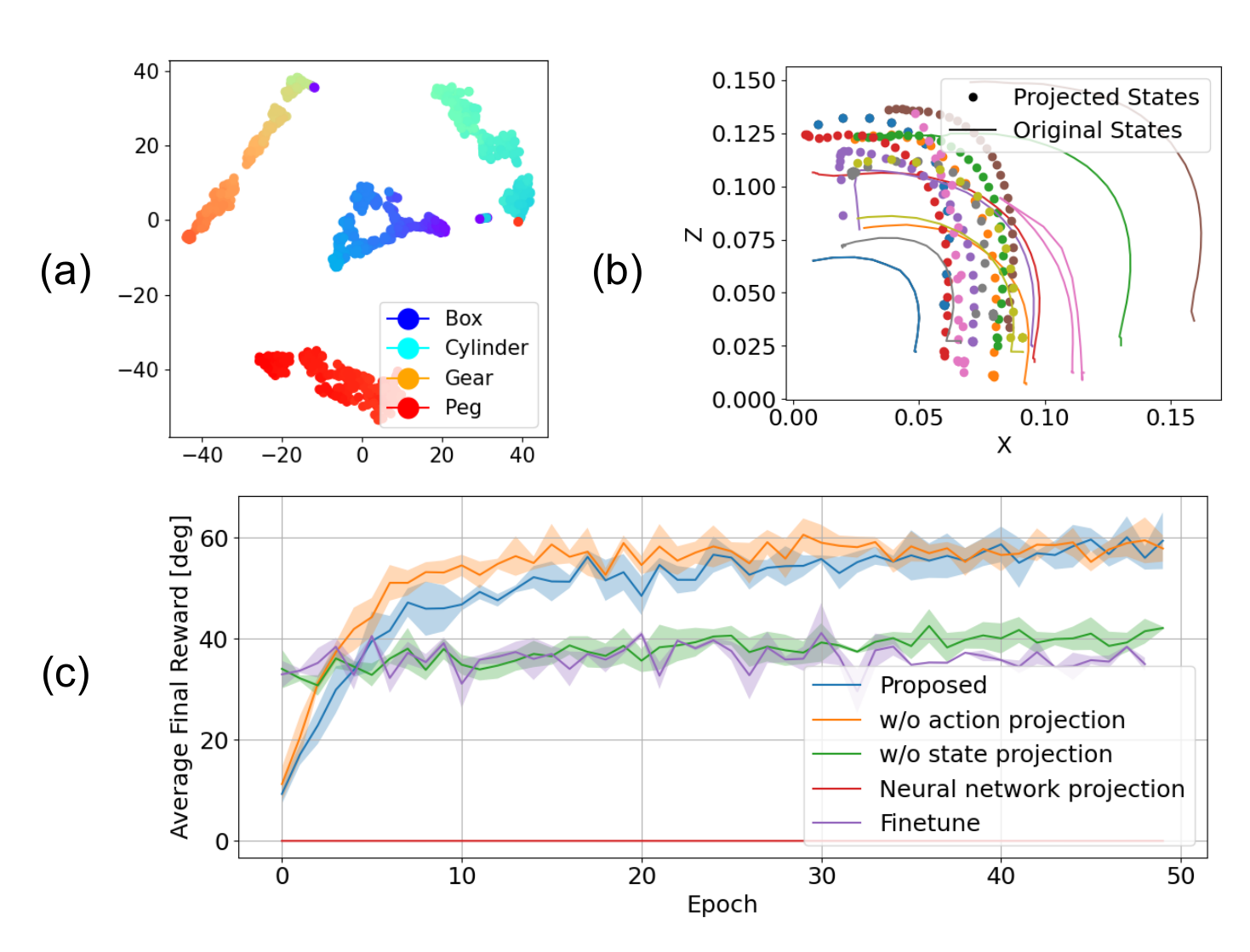}
    \caption{a) t-SNE visualization of learned object features, color difference within the same class indicates different object size, b) pivoting trajectories before and after projection of different objects, c) learning curves for policy adaption}
    \label{fig:LC}\vspace{-5mm}
\end{figure}

\begin{figure*}
    \centering
    \includegraphics[width=370pt]{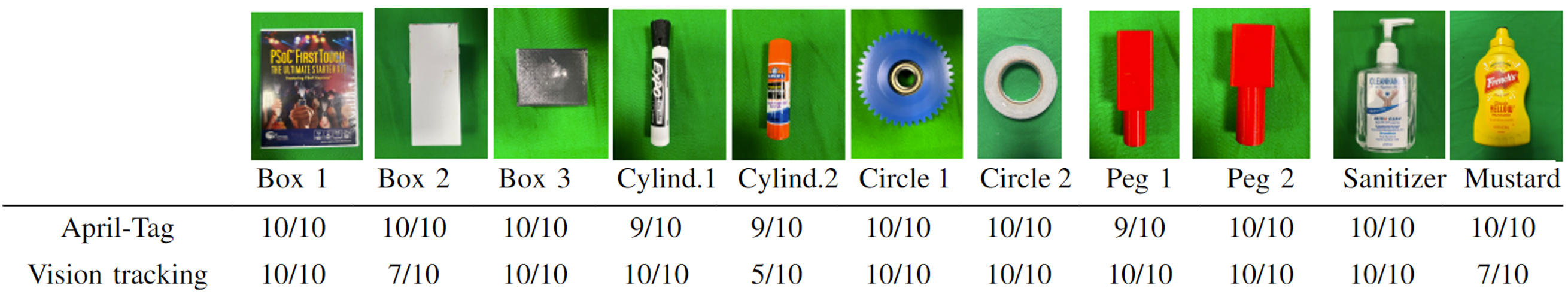}
    \caption{Test objects and success rates for the two vision systems}
    \label{fig:realworld_success_rate}
\end{figure*}

\textbf{Representation Learning of Object Features:} The encoding neural network is trained in a supervised fashion using the dataset $\mathcal{D}$ described in Section~\ref{subsec:representation}, and the computed object feature space is evaluated using the t-SNE method \cite{van2008visualizing}. As shown in Fig~\ref{fig:LC}(a), object features are clustered into four groups in the learned feature space, representing four object classes. In addition, within each class, this feature space can distinguish the size information of different objects, which shows the learned feature space can be utilized for downstream tasks.

\textbf{State/Action Projection Nets Training:}
Once the unitary policy and the object features are learned, the \emph{State/Action Projection Nets} are trained in the same simulation environment (see Section~\ref{sec:approach}) by selecting 40 randomly-sized objects, 10 objects for each class $c_{obj}$. The two projection nets are two-layer Relu networks with 16 and 35 units, respectively, and batches of $200$ trajectories are selected to update networks for each epoch of the algorithm. Since the elements of quaternions in the state are coupled and will be distorted by the transformation, we only apply the same projection to both the object and gripper positions for the state projection net, that is $T_s \in \R^{3\times 3}$ and $T_a\in \R^{4\times 4}$ remains unchanged. 

\textbf{Multi-objects generalization performances:} The proposed approach is evaluated in simulation and compared against three ablation studies and two baselines: 
\begin{enumerate}
    \item \emph{NN projections}: train two neural networks $\bar{s}_u = \rho_s (f,s)$, $a = \rho_a (f,\bar{s}_u)$ to replace the linear transformations in the state/action projection nets;
    \item \emph{w/o state}: our approach without state projection net;
    \item \emph{w/o action}: our approach without action projection net;
    \item \emph{Finetune}: the unitary policy is fine tuned using PPO without state and action projection nets;
    \item \emph{Pearl}: train one s.o.t.a. Meta-RL approach, Pearl \cite{rakelly2019efficient}.
\end{enumerate}
Approaches 1) to 4) are trained following Alg.~\ref{algo} but updating different networks. PEARL is learned from scratch.

As depicted in Fig~\ref{fig:LC}(c), the proposed approach and the ablation \emph{w/o action} outperform all the other approaches. The \emph{NN projections} does not adapt the unitary policy to multi-objects, possibly because it cannot use the structure of the linear transformation and might require much more data to learn the task. We also notice that the state projection net helps the most for adaption and converges faster than the proposed method. The reason is that the proposed method's action projection is not perfect initially and slows down the training. However, as shown in Table~\ref{tab:sim_success_rate}, the proposed method achieves a higher success rate than all the baselines, which indicates the action projection helps to adjust policy according to the object feature. 
In our experiments, Pearl can approach the objects but cannot learn to pivot for multiple objects. We conclude that the objects are too different, making it difficult for Pearl to learn a policy for all objects from scratch. Since the objects in the simulation are randomly generated, some objects have small surfaces to stand on compared to their sizes, which makes them challenging to pivot, and our approach fails.

\begin{table}[http]
    \centering
    \begin{tabular}{c|c|c|c|c|c}
         Ours & w/o action &w/o state &Finetune&NN&Pearl\\
         \cline{1-6}
         $\bm{30/40}$& 27/40 & 19/40 &15/40 & 0/40&0/40\\
    \end{tabular}
    \caption{Average success rates on 40 objects in the training environment over 3 random runs}
    \label{tab:sim_success_rate}
\end{table}
We further analyze the effect of the state projection net by plotting the pivoting trajectories of different objects before and after the projection in the two most representative axis $X,Z$. As depicted in Fig~\ref{fig:LC}(b), even though the original trajectories are very diverse (solid lines), the projection net brings the pivoting trajectories together, enabling the unitary policy to adapt to new objects.

\subsection{Real-World Experiments}
\begin{figure}
    \centering
    \includegraphics[width=180pt]{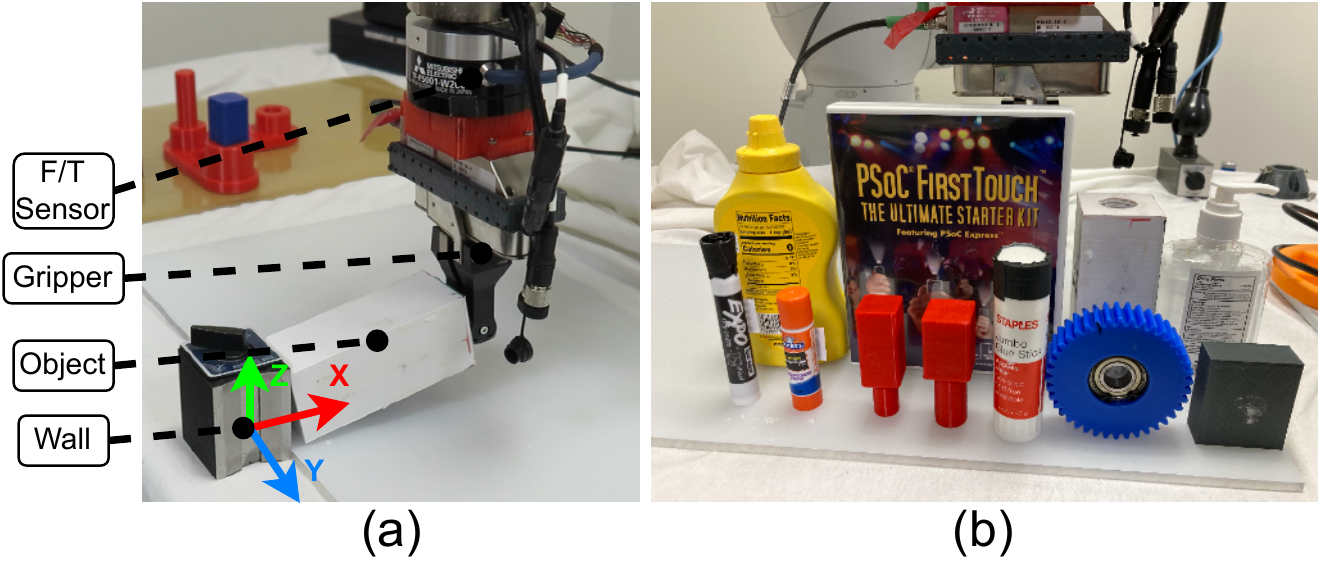}
    \caption{a) Real-world setup. b) Objects in the real world. }
    \label{fig:realworld}
\vspace{-0.7cm}
\end{figure}

\textbf{Real Robot Setup and Vision Feedback:} We use a 6DoF Mitsubishi Assista RV-5AS-D collaborative robot arm with a WSG32 gripper as shown in Fig.~\ref{fig:realworld}(a). The robot is controlled in impedance control mode with stiffness set to 12 $N/mm$. The external forces are measured by wrist mounted F/T sensor. For state estimation, we compare two systems to obtain the object pose from an RGB-D camera (Intel Realsense D435). The first system uses April-tags \cite{olson2011apriltag} on the objects for tracking. This system provides high-accuracy state information but needs the tag's placement. The second system comprises vision-based 6D pose estimation consisting of a mask and deep feature extraction~\cite{mayer2022transforming,revaud2019r2d2} pipeline with pose tracking. To accomplish fast tracking of novel objects in motion using RGB-D images, we introduce several augmentations to enhance the pose tracking~\cite{earl} based on the BundleTrack~\cite{wen2021bundletrack}. The method does not require CAD models, and because of sensor noise, the state estimation can be noisier. We test pivoting manipulation with both systems to evaluate the robustness of our proposed approach.

\textbf{Object Dataset:} Fig~\ref{fig:realworld}(b) shows the objects we used for the real-world experiments. We considered nine objects which can be categorized into four object classes in the simulation to test the sim-to-real transfer performance of the proposed approach. Furthermore, we test the generalizability on two irregularly shaped objects (sanitizer and mustard bottle). Please note that none of these objects was seen during the simulation and the same hold for depth images of the real objects taken by the camera and used to infer object features.

\textbf{Sim-to-Real Transfer and Generalization of Pivoting:} The proposed approach is evaluated with zero-shot transfer learning over all objects. An experiment is deemed successful when an object reaches a stable stand-up position, and the success rates are shown in Fig~\ref{fig:realworld_success_rate}. We start the analysis considering the April-Tag system to estimate the state. Even though the shapes and sizes of the objects are very different, i.e., $l_x=[6,18.5], l_y=[1,15], l_z=[1,5] [cm]$. 
our approach achieves $100\%$ success rate on almost all the objects, demonstrating direct sim-to-real transfer capability and generalization to multiple objects. The failure cases happen all in objects with a cylindrical base, i.e., Cylinder 1,2 and Peg 2, because their shape is prone to a rolling behaviour and have a smaller base to stand on. However, the policy is robust to these difficulties in most of the cases and failed only once.
Successful pivoting experiments are visualized with a sequence of snapshots in Fig~\ref{fig:snapshots}(b)-(c). Noticeably, the learned policy can recover after failure has shown in Fig~\ref{fig:snapshots}(c). The first pivoting attempt fails probably because the box is the heaviest object we have ($450 [g]$ while the other objects range between $[0,200] [g]$), and this might enhance a slipping effect in the contact when pushing. However, the learned policy moves behind the object and applies a successful pivoting the second time, showing compelling robustness properties.

Next, we test the generalization to two out-of-distribution objects: ``sanitizer'' and the ``mustard bottle''. The shapes of these two objects are complex, non-convex and with irregular contact surfaces that are not flat. Moreover, no similarly shaped object is considered in training. However, our approach succeeds on these two objects with $100\%$ success rate and demonstrates generalizability properties.

Finally, we evaluate the performance when using the vision tracking system that returns more noisy state estimations. As shown in Fig~\ref{fig:realworld_success_rate}, we still achieves $100\%$ success rates for most objects. The failure cases are due to excessive noise in the state: in Box 2 because the object has no texture, in Cylinder 2 because the object is very small, and for the Mustard bottle the tracking gets lost when close to the goal.

\section{CONCLUSIONS}
We propose a framework to learn robotic skills for pivoting real-world objects. The method is trained only in simulation, requiring only one depth image of the manipulated object to transfer to real-world tasks. Moreover, the same policy generalizes to pivot multiple real-world objects. The main idea is based on learning a robust RL policy for a ``unitary'' object and then learning two projection networks that adapt the states and actions fed into/outputted by such a policy. An object feature space is learned from top-down view depth images of the objects to encode the kinematic properties such as size and shape. The real-world experiments show a successful zero-shot transferring for sim2real gap and generalization to multiple objects. 

The proposed approach adapts the state/action space based on object visual features. However, other characteristics, such as friction and inertial properties, can also be considered to generalize our approach to a broader class of tasks using additional sensor inputs or system-ID procedures.  Furthermore, we apply linear transformations to the states and actions to have more structured models and gain data efficiency. Still, more complex manipulation tasks require learning state-dependent nonlinear transformations.



\bibliographystyle{IEEEtran}
\bibliography{IEEEabrv1}

\end{document}